\documentclass[conference]{IEEEtran}
\IEEEoverridecommandlockouts
\usepackage{cite}
\usepackage{amsmath,amssymb,amsfonts}
\usepackage{algorithmic}
\usepackage{graphicx}
\usepackage{textcomp}
\usepackage{xcolor}

\usepackage{url}
\usepackage[utf8]{inputenc}
\usepackage{booktabs}
\usepackage{balance}

\usepackage{subfig}
\usepackage{xspace}
\usepackage[linesnumbered,ruled,vlined]{algorithm2e}

\newcommand{\method}{\textsc{PoisonProbe}\xspace}

\newtheorem{definition}{Definition}
\DeclareMathOperator\arctanh{arctanh}

\def\BibTeX{{\rm B\kern-.05em{\sc i\kern-.025em b}\kern-.08em
    T\kern-.1667em\lower.7ex\hbox{E}\kern-.125emX}}
\begin{document}

\title{Indirect Adversarial Attacks via Poisoning Neighbors for Graph Convolutional Networks$^*$\thanks{$^*$This paper is the full version of \cite{takahashi2019indirect}.}}

\author{\IEEEauthorblockN{Tsubasa Takahashi}
\IEEEauthorblockA{
\textit{LINE Corporation}\\
%Tokyo, Japan \\
tsubasa.takahashi@linecorp.com}
}

\maketitle

\begin{abstract}
Graph convolutional neural networks, which learn aggregations over neighbor nodes, have achieved great performance in node classification tasks.
However, recent studies reported that such graph convolutional node classifier can be deceived by adversarial perturbations on graphs.
Abusing graph convolutions, a node's classification result can be influenced by poisoning its neighbors.
Given an attributed graph and a node classifier, how can we evaluate robustness against such indirect adversarial attacks?
Can we generate strong adversarial perturbations which are effective on not only one-hop neighbors, but more far from the target?
In this paper, we demonstrate that the node classifier can be deceived with high-confidence by poisoning just a single node even two-hops or more far from the target.
Towards achieving the attack, we propose a new approach which searches smaller perturbations on just a single node far from the target.
In our experiments, our proposed method shows 99\% attack success rate within two-hops from the target in two datasets.
We also demonstrate that $m$-layer graph convolutional neural networks have chance to be deceived by our indirect attack within $m$-hop neighbors.
The proposed attack can be used as a benchmark in future defense attempts to develop graph convolutional neural networks with having adversary robustness.
\end{abstract}


\section{Introduction}

Graph is a core component for many important applications ranging from recommendations and customer type analysis in social networks to anomaly detection, behavior analysis in sensor networks \cite{yoon2019fast} \cite{gao2018graph}. 
Even if graph is not explicitly given, estimated latent graph can be helpful for those applications because the graph gives them relationships and interactions between nodes \cite{hallac2017toeplitz} \cite{takahashi2017autocyclone} \cite{kipf2018neural}.
One of the most frequently applied tasks on graph data is node classification: given a single large attributed graph and the class labels of subset of nodes in the graph, how to predict the labels of the remaining nodes.

The last years, deep neural networks for large graphs have achieved great performance in node classification problems \cite{kipf2017semi} \cite{hamilton2017inductive} \cite{ying2018graph}.
One of the well-known approaches in node classification is graph convolutional neural networks (GCNs).
GCNs utilize not only node features, but relational information on graph to perform classification task.

On the other hand, recent years many researchers noticed that deep learning architectures can easily be fooled \cite{szegedy2014intriguing}.
Even only slight, deliberate perturbations, it can lead a machine learning model to misclassification \cite{carlini2017towards} \cite{athalye2018obfuscated}.
The perturbation is called \textit{adversarial perturbation} and a sample with the perturbation is known to \textit{adversarial example}.
The adversarial examples is a potentially critical safety issues in any machine learning based systems.
Therefore studies about generating adversarial examples are important to evaluate robustness of the target machine learning models \cite{goodfellow2014explaining} \cite{sharif2016accessorize} \cite{eykholt2017robust} \cite{kakizaki2019glassmasq}.
As well as typical deep learning architectures, GCNs are also highly vulnerable in adversarial perturbations \cite{zugner2018adversarial}.

\begin{figure}
    \centering
    \includegraphics[width=.9\hsize]{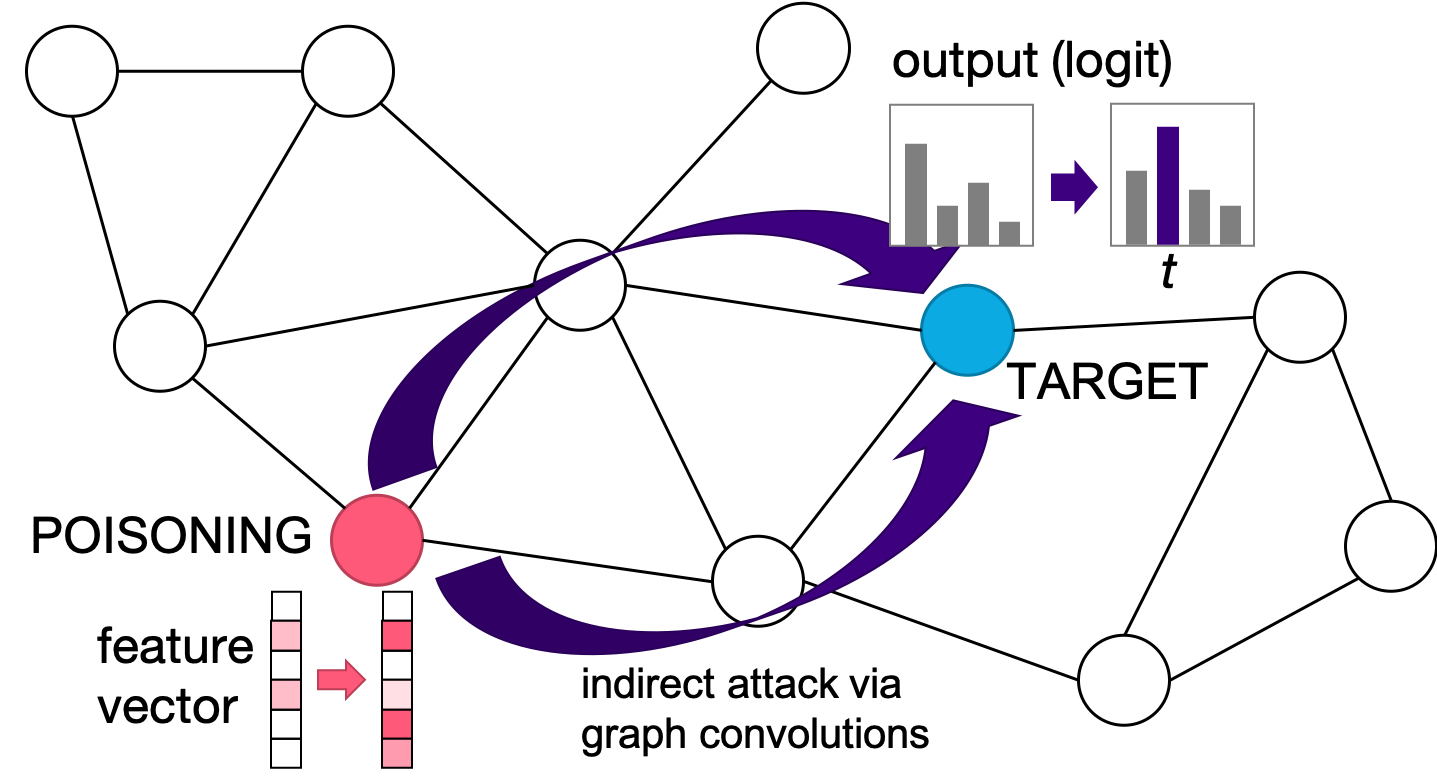}
    \caption{Indirect adversarial attack by poisoning a single node. Poisoned information is propagated through graph and influences other nodes' classification results.}
\end{figure}

Against typical neural networks, adversarial perturbations for GCNs have several particular characteristics.
First, we can add perturbations on both features and edges. 
Second, we can lead misclassification by not only direct perturbations on the target, but indirect perturbations on the target's neighbors.
\cite{zugner2018adversarial} proposed an adversarial perturbation method to perform both direct and indirect attacks for GCNs in semi-supervised learning setting.
The indirect attack iteratively perturbs either a feature or an edge for given number of 1-hop neighbors.

Additionally, we consider about another possibility of adversarial examples on graphs.
A series of graph convolutions delivers nodes' information through series of edges.
Thus, even if there is no direct connection, the poisoned information possibly influence a node far from the poisoned node.
Hence, not only around directly connected neighbors, but we also consider about adversary robustness against such indirect attacks from remote nodes.
Further, to evaluate the upper bound of the robustness of GCNs, we need an attack which is sufficiently strong to deceive them.
A strong attack against GCNs is perturbations on few nodes far from the target.
To evaluate the robustness of node classifiers with graph convolutions, we need a method to generate such strong indirect perturbations.

In this work, we attempt to close the gaps.
The question we want to solve is:
given an attributed graph and its node classifier with GCN layers,
how can we craft high-confidence adversarial perturbation which leads misclassification into a target node thorough poisoning a single node far from the target?
Towards evaluating adversary robustness of GCNs, the problem addressed here has significant importance.

\noindent
\textbf{Present Work.}
To answer these questions, we introduce an adversarial perturbation method \method which poisons just a single node's features to lead misclassification into a target node far more than one-hop from the poisoned node.
The proposed method enable us to evaluate GCNs' robustness against indirect adversarial perturbations.

\noindent
\textbf{Contributions.}
This paper makes the following contributions:
\begin{enumerate}
    \item We introduce a new attack named \method which deceive node classifiers with GCN. \method poisons a node's features to lead misclassification into a target far more than 1-hop from the poisoned node. 
    \item We also introduce an approach to find the poisoning node that has high chance to result in the smallest perturbations than other candidates.
    \item Our proposed attack is significantly more effective than previous approach. In our experiments, proposed method with poisoning randomly selected node shows at least 92\% attack success rate within two-hops from the target for two datasets. Further, the proposed method with the poisoning node selection shows 99\% attack success rate at two-hops.
    \item We reveal that $m$-layer GCNs have chance to be deceived by our attack within $m$-hops from the target.
\end{enumerate}
The proposed attack can be used as a benchmark in future defense attempts to develop graph convolutional neural networks with having adversary robustness.

\section{Related Work}

\noindent
\textbf{Deep Learning for Graphs.} 
Researches of deep learning for graphs can be distinguished in two parts: node embeddings \cite{grover2016node2vec} \cite{perozzi2014deepwalk}
\cite{bojchevski2018deep} and graph neural networks \cite{kipf2017semi} \cite{kipf2018neural} \cite{hamilton2017inductive}.
We focus on the latter, especially graph convolutional neural networks (GCNs) and adversarial attacks for them.

While many classical approaches have been introduced in the past, the last years, deep neural networks for large graphs have achieved great performance in node classification problems \cite{ying2018graph}.
The core idea behind GCNs is to learn how to aggregate feature information over local graph neighborhoods using neural networks.
A single graph convolution operation aggregates feature information over a node’s one-hop neighbors on a graph, and by stacking multiple those convoluted information can be propagated through the graph.

Under such GCNs, as well as typical deep learning architectures, GCNs are also vulnerable in adversarial perturbations \cite{zugner2018adversarial}.
This paper also tackles crafting adversarial perturbations under GCNs, but focus on indirect perturbations on a node far from a target node.

\noindent
\textbf{Adversarial Examples.} 
Adversarial example is a crafted input for deceiving deep neural networks \cite{szegedy2014intriguing}.
It is a potentially critical safety issues in machine learning based systems.
Therefore, we need to evaluate adversary robustness at developing machine learning models.
One of simple defense approach against adversarial examples is to mask gradients \cite{papernot2016distillation}.
However, it provides a false sense of security \cite{carlini2017towards} \cite{athalye2018obfuscated}. 
Adversarial training \cite{madry2017towards}, which injects adversarial examples with correct labels into training samples, is a simple way to re-train a model.
Recently, several certified robust learning approaches have been proposed \cite{wong2018provable} \cite{mirman2018differentiable} \cite{tsuzuku2018lipschitz}.
The certified defense mechanisms practically do not have enough robustness at all, but give certifiable robustness around training points.

On the other hands, studies about generating adversarial examples / perturbations are important to evaluate robustness of the target machine learning models \cite{sharif2016accessorize} \cite{eykholt2017robust} \cite{kakizaki2019glassmasq} \cite{yakura2019robust}.
The most well known method is FGSM (Fast Gradient Sign Method) \cite{goodfellow2014explaining}.
FGSM finds adversarial perturbations optimized for the $L_{\infty}$ distance metric.
FGSM is a light-weight method to craft adversarial perturbations.
Thus, it is a standard way to try finding adversarial examples.
\cite{kurakin2016adversarial} proposed an iterative approach extending FGSM.
Deepfool is an untargeted attack optimized for the $L_2$ distance metric in an efficient way.
CW Attack \cite{carlini2017towards} discovers an adversarial perturbation with small size of perturbations.
CW Attack has variants for the $L_{0}$, $L_{2}$ and $L_{\infty}$ distance metrics.
This paper we assume $L_{2}$ for CW Attack.
CW Attack is used as a standard benchmark of adversary robustness of neural networks.

\noindent
\textbf{Adversarial Perturbations on Graphs.}
Works on adversarial attacks for graph learning tasks are only a few works. 
\cite{dai2018adversarial} introduced an adversarial attack which exploits reinforcement learning ideas through deleting edges. 
\cite{zugner2019adversarial} proposed untargeted attack for graph via meta learning.
\cite{zugner2018adversarial} proposed \textit{Nettack} which is a crafting adversarial perturbation on graph to deceive its node classifier through perturbing both features and edges.
Nettack can deceive a node's classification result in two ways: direct and indirect.
Nettack's direct attack perturbs both features and edges of the target node.
While indirect attack called \textit{influencer attack} picks the target's 1-hop neighbors (called influencer nodes), then iteratively perturbs either a feature or an edge of influencer nodes with in budget of perturbations.
Nettack prefers to choose a feature to perturb in early iterations, because adding/removing edge can close to successful adversarial example than perturbing a single feature.
The one by one perturbations for features is not powerful to find successful adversarial examples.
On the other hand, the perturbations of Nettack seem data poisonings on a graph aiming to deceive node classifiers.
\cite{zhu2019robust} proposed robust graph convolutional networks (RGCN) against adversarial attacks.

\noindent
\textbf{Difference against Existing Works.}
In this paper we mainly study on adversarial perturbations on a single node far from the target more than one-hop, whereas Nettack perturbs multiple one-hop neighbors for indirect attacks.
Since typical GCNs have two-layers of graph convolutions, there is chance to deliver poisoned feature information from two-hops and more.
Our goal is to clarify a potential vulnerability of GCNs, and make more strong attack for evaluating adversary robustness of GCNs.
Such strong attacks are important to measure the robustness of GCNs including RGCN \cite{zhu2019robust}.
We also assume no changes on graph structure.
We have many applications assuming stable graph. 
Sensor network is one of those applications.
We here study about vulnerability of GCNs under the assumption of no structural changes.
Further, Our study do not explicitly assume the perturbation budget which Nettack introduced, but the proposed method can perform like having the budget by rejecting the results over the budget.

\section{Preliminary}
\label{sec:preliminary}

We consider the task of (semi-supervised) node classification in a single large graph having binary node features.
Formally, let $G = (X,A)$ be an attributed graph, where $A\in \mathbb{R}^{n \times n}$ is the adjacency matrix representing the connections and $X \in \mathbb{R}^{n \times d}$ represents the nodes’ features.

\paragraph{Node Classification with GCNs.}
\label{sec:nodeclass}
GCN (Graph Convolutional Neural Network) is a semi-supervised learning method to classify nodes, 
given feature matrix $X$, adjacency matrix $A$ and labels for subset of nodes in the graph.
We have several variants of GCN, but we assume the most common way \cite{kipf2017semi}. 
The GCN is defined as follows:
\begin{equation*}
    H^{(l+1)} = \sigma (\tilde{D}^{-\frac{1}{2}} \tilde{A} \tilde{D}^{-\frac{1}{2}} H^{(l)} W^{(l)})
\end{equation*}
where $\tilde{A}=A+I_N$, $I_N$ is the identity matrix, $\tilde{D}_{ii}=\sum_j \tilde{A}_{ij}$, $H^{(0)}=X$ and $\sigma$ is an activation function.
We assume $\sigma(x)=ReLU(x)$.
Here, let $f(X,A)$ be the output of neural network $f$ with Softmax layer, and we denote logits $Z(X,A)$ that is the output before Softmax layer as
    $f(X,A) = Softmax(Z(X,A))$.
A commonly used application is two-layer GCN.
That is, 
    $Z(X,A) = \hat{A}\sigma(\hat{A}XW^{(0)})W^{(1)}$
, where $\hat{A}=\tilde{D}^{-\frac{1}{2}} \tilde{A} \tilde{D}^{-\frac{1}{2}}$. 

\paragraph{Adversarial Examples against GCNs.}
We introduce adversarial examples on graph against GCNs.
Let \textbf{positive adversarial example} be an adversarial example which satisfies the following condition:
\begin{equation}
    \max_{i \neq t}(Z(X', A)_{u,i}) - Z(X',A)_{u,t} < 0
    \label{eq:target}
\end{equation}
where $Z(X', A)_{u,i}$ is $u$' logit value of class $i$.
Positive adversarial examples cause misclassification into the target node $u$ to be the target class $t$.
Untargeted attack is also described as:
\begin{equation}
    Z(X',A)_{u,c} - \max_{i \neq c}(Z(X', A)_{u,i}) < 0
    \label{eq:untarget}
\end{equation}
where $c$ is $u$'s legitimate output $f(X,A)_u$.

\paragraph{Box Constraints.}
We ensure the modification by adversarial perturbations yields a valid input, we have a constraint on modification $\delta$ as $0 \le x+\delta \le 1$.
\cite{carlini2017towards} introduced following change-of-variables:
\begin{equation}
    x + \delta = \frac{1}{2}(\tanh(w)+1).
\label{eq:changeval}
\end{equation}
Since $-1 \le \tanh(w) \le 1$, it follows that $0 \le x+\delta \le 1$.
In the above notation, we can optimize over $w$ to find valid solution with a smoothing effect of clipped gradient descent that eliminates the problem of getting stuck in extreme regions.
This method allows us to use an optimization algorithm that does not natively support box constraints.
We also employ Adam optimizer \cite{kingma2014adam} with this box constraints in our attack.
\section{Proposed Node Poisoning}

Given the node classification setting described in section \ref{sec:preliminary}, our goal is to find small perturbations on features of a node on a graph $G=(X,A)$ and simultaneously to lead misclassification into the other node even when those two nodes have no direct connection.
Hence, we assume $u \neq v$, where $u$ is the target node and $v$ is the node which we add the perturbations.
We also assume no structural changes on graph.

To solve the problem above, an optimization based approach is introduced in section \ref{sec:optim}.
We also introduce \textit{poisoning node selection} that discovers the poisoning node that has high  chance to result in the smallest perturbations than other candidates (section \ref{sec:pns}).

\subsection{\method}
\label{sec:optim}

Here we propose a new attack \method that solves the above problem.
The objective function \method solves is defined as
\begin{equation}
    \mathcal{L}(x'_v, u, t, \lambda) = ||x_v - x'_v||_2 + \lambda \cdot g(X', u, t)
    \label{eq:L}
\end{equation}
\begin{equation}
    g(X', u, t) = \left(\max_{i \neq t}(Z(X', A)_{u,i}) - Z(X',A)_{u,t}\right)_+
\label{eq:g}
\end{equation}
where $x_v$ is feature vector of $v$, $x'_v = x_v + \delta_v$, $\delta_v$ represents perturbations on $x_v$, $t$ is the target class label, $X'=X+ e_v^T \cdot \delta_v$, $e_v\in \{0,1\}^{n}$ is an unit vector that $v$-th element is 1 and $(\cdot)_+=\max(\cdot,0)$.
We then try to solve the following optimization problem to find the high-confidence adversarial perturbations on $v$ targeting $u$.
\begin{equation*}
    x^*_v = \arg \min_{x'_v} \mathcal{L}(x'_v, u, t, \lambda)
\end{equation*}

In the optimization, we employ the transformation (\ref{eq:changeval}) for $x'_v$ to find valid solution with a smoothing effect of clipped gradient descent that eliminates the problem of getting stuck in extreme regions.
To find indirect adversarial perturbations, we need to estimate gradient of $\mathcal{L}(w_v, u, t, \lambda)$ over $w_v$ which is a transformed variable from $x'_v=x_v+\delta_v$.
In case the connection between $u$ and $v$ is indirect, perturbation on $v$ is propagated via graph convolution into $v$ with damping.
Hence, it is not easy to find the solution that turn $u$'s output into the targeted class.
To find the valid solution satisfying (\ref{eq:g}), we employ binary search for discovering effective value of $\lambda$.

The detail algorithm of \method is described in Algorithm \ref{alg1}.
The proposed algorithm consists of outer loop and inner loop.
The inner loop discovers smaller perturbations which can achieve desired targeted attack under the several parameters are fixed.
Then, the outer loop discovers the parameters iteratively by utilizing binary search.
If the inner loop successfully find the adversarial perturbation, it decrease the constant $\lambda$.
While, if the inner loop cannot find it, it increase $\lambda$, that means it focuses on finding perturbations satisfying (\ref{eq:g}) rather than reducing size of perturbations next.

\SetKwInOut{Parameter}{Parameter}

\begin{algorithm}[t]
\caption{\method}         
\label{alg1}
\DontPrintSemicolon
\KwIn{feature matrix $X$, adjacency matrix $A$, target node $u$, target class $t$, poisoning node $v$}
\KwOut{adversarial example $x^*_v$}
\Parameter{$\lambda_{init}$, $\lambda_{min\_init}$, $\lambda_{max\_init}$, learning rate $\gamma$, max\_search\_steps, max\_iter}
$\lambda \leftarrow \lambda_{init}$;
$\lambda_{max} \leftarrow \lambda_{max\_init}$;
$\lambda_{min} \leftarrow \lambda_{min\_init}$\;
$\Tilde{x}_v \leftarrow \arctanh(x_v)$\;
min\_dist $\leftarrow \infty$\;
\For{$step=1$ to max\_search\_steps}{
    is\_found\_adversarial $\leftarrow$ false \;
    \For{$i=1$ to  max\_iter}{  
        ${w_v} \leftarrow {w_v} - \gamma \nabla_{w_v} \mathcal{L}(w_v, u, t, \lambda)$\;
        $x'_v \leftarrow \frac{1}{2}(\tanh(\Tilde{x}_v+{w_v})+1)$\;
        dist $\leftarrow ||x_v-x'_v||_2$\;
        \If{$g(X', u, t) < 0$}{
            \If{dist $<$ min\_dist}{
                $x^*_v \leftarrow x'_v$\;
                min\_dist $\leftarrow$ dist\;
            }
        is\_found\_adversarial $\leftarrow$ true\;
        }
    }
    \uIf{$is\_found\_adversarial$}{
        $\lambda_{max} \leftarrow \lambda$\;
    }\Else{
        $\lambda_{min} \leftarrow \lambda$\;
    }
    $\lambda \leftarrow \frac{1}{2}(\lambda_{min}+\lambda_{max})$\;
}
\Return{$x^*_v$}

\end{algorithm}

\subsection{Poisoning Node Selection}
\label{sec:pns}

Here we discuss about how to choose the poisoning nodes to achieve the misclassification with smaller perturbations.
On a graph convolutional neural network, a graph convolution layer aggregates features of both a (center) node and its 1-hop neighbors.
Here we assume every 1-hop neighbor can equally deliver its features to the center node $u$ through graph convolution.

\begin{definition}(Poisoning efficiency of 1-hop neighbor)
Let $\mathcal{N}_u^{(1)}$ be the set of 1-hop neighbors around $u$.
Poisoning efficiency of $v \in \mathcal{N}_u^{(1)}$ to lead misclassification towards $u$ is defined by
\begin{equation}
    \phi_u^{(1)}(v) = \frac{1}{|\mathcal{N}_u^{(1)}|} .
    \label{eq:1hef}
\end{equation}
\end{definition}

Next, we consider about the poisoning efficiency of a node far more than 1-hop.
To simplify, we only consider shortest paths that are paths ignoring edges between nodes where are the same distance from the target.
This simplification enables us to transform the graph into the tree whose root is the target $u$.
We call the tree {\it neighborhood tree}.

\begin{figure}[t!]
    \centering
    \subfloat[Poisoning at 2-hop neighbors]{
        \includegraphics[width=.485\hsize]{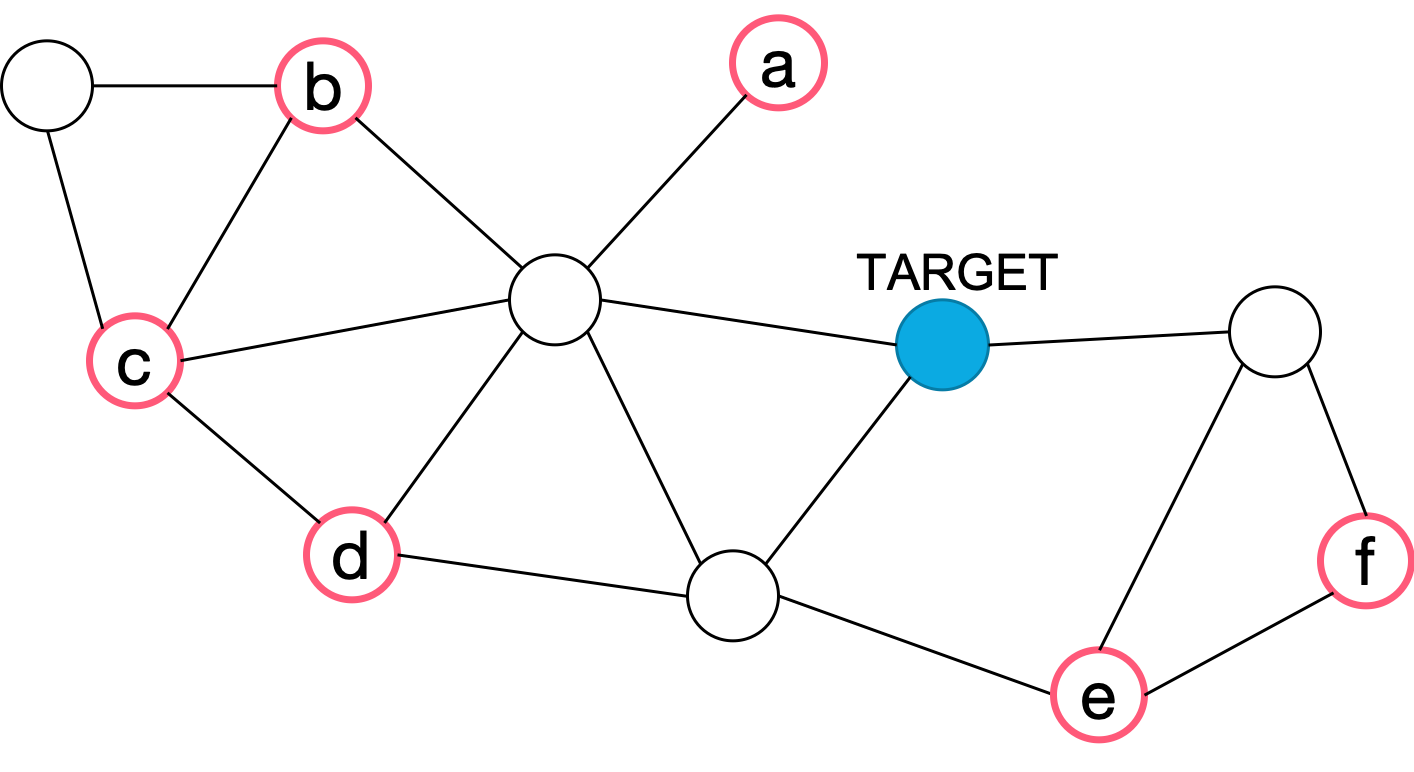}
        \label{fig:cand_graph}
    }
    \subfloat[Neighborhood Tree]{
        \includegraphics[width=.485\hsize]{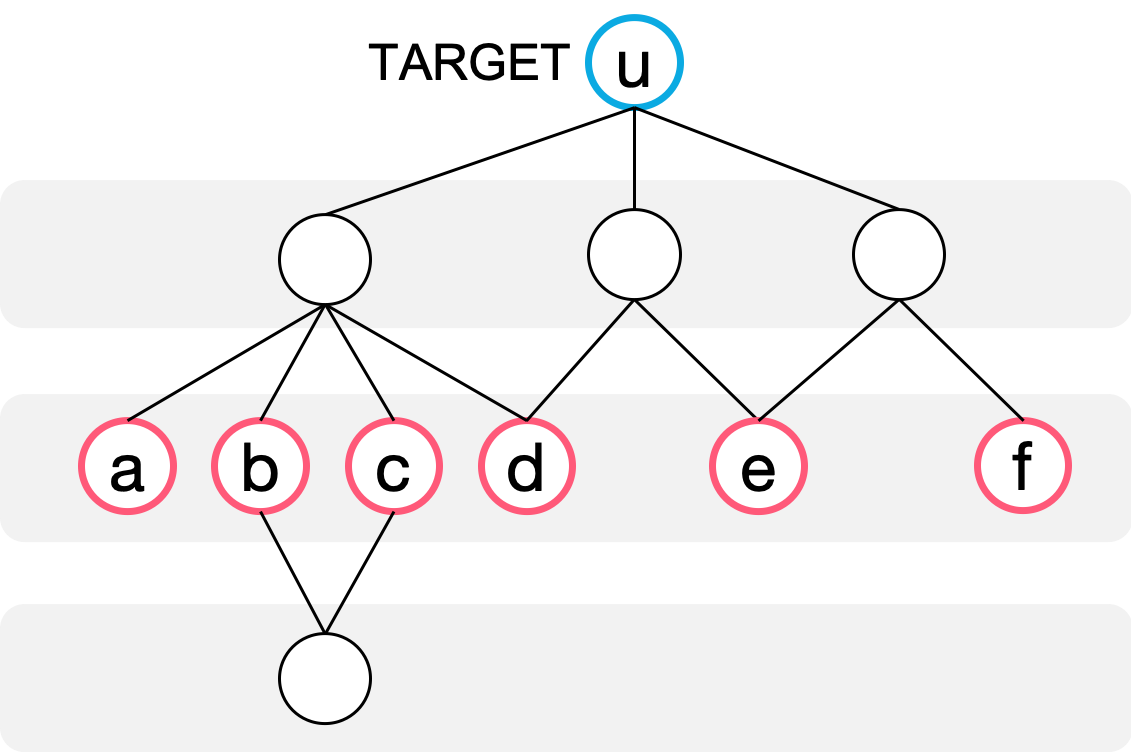}
        \label{fig:ntree}
    }
    \caption{Transformation from graph to neighbor tree.
    We want to choose the most efficient poisoning node from the 2-hop neighbors (pink) of the target (blue). We transform the graph (a) into the neighbor tree (b) to easily discover ancestor of candidates nodes and shortest paths towards the target.}
    \label{fig:infl_overview}
\end{figure}

\begin{definition}(Poisoning efficiency of $m$-hop neighbor)
Let $\mathcal{N}_u^{(m)}$ be the set of $m$-hop neighbors around $u$ and $anc_u(v)$ be the ancestor node of $v$ in the neighborhood tree whose root is $u$.
Poisoning efficiency of $v \in \mathcal{N}_u^{(m)} (m>1)$ to lead misclassification towards $u$ is defined as
\begin{equation}
    \phi_u^{(m)}(v) = \sum_{\eta \in anc_u(v)} \frac{1}{|\mathcal{N}_\eta^{(1)}|} \phi_u^{(m-1)}(\eta) 
    \label{eq:mhef}
\end{equation}
where
\begin{equation}
    \phi_u^{(1)}(\eta) = 1 .
    \label{eq:1hef2}
\end{equation}
\end{definition}

Actually (\ref{eq:1hef}) is better interpretation of poisoning efficiency of a 1-hop neighbor than (\ref{eq:1hef2}).
However we utilize (\ref{eq:1hef2}) for poisoning efficiency of a $m$-hop neighbor because (\ref{eq:1hef}) is the equivalent for all 1-hop neighbors.

The score of poisoning efficiency defined the above looks like band-width of the path delivering poisoned information from the candidate to the target.
If the score is small, the poisoned information will be shrunk through the path.
Therefore, we need to enlarge the poisoned information to achieve the adversarial attack.
While if we can select the path whose poisoning efficiency is high, we have chance to reduce the amount of perturbations to do it.

From $u$'s $m$-hop neighbors $\mathcal{N}_u^{(m)}$, we pick a node  which has the maximum poisoning efficiency:
\begin{equation}
    v^* = \arg \max_{v \in \mathcal{N}_u^{(m)}} \phi_u^{(m)}(v) .
\end{equation}

In case there is multiple nodes having the maximum poisoning efficiency, we randomly select one node from them.
For the 1-hop neighbors, we always randomly pick a node from $\mathcal{N}_u^{(1)}$ because all the candidates share the same poisoning efficiency.

\section{Extensions}

We here describe simple extensions of \method.

\subsection{Extension 1: Multiple Node Perturbation}
\label{sec:ext2}

We introduce an extension to perturb multiple nodes.
We here describe the difference fom the original \method.

\subsubsection{Targeted Perturbation}

The formulation of the problem we solve here is described as follows:
\begin{equation*}
    \text{minimize } ||X_V - X'_V||_2
\label{eq:obj2}
\end{equation*}
\begin{equation*}
    \text{such that } \arg \max(Z(X',A)_{u}) = t
\end{equation*}
where $X_V$ is feature matrix of $V$.
$V$ includes poisoning nodes $\{v_1, \dots, v_{|V|}\}$.
The loss function (\ref{eq:L}) is rewritten as:
\begin{equation*}
    \mathcal{L}(X'_V, u, t, \lambda) = ||X_V - X'_V||_2 + \lambda \cdot g(X', u, t).
    \label{eq:L2}
\end{equation*}
We can also employ Algorithm (\ref{alg1}) to discover perturbations on $\{v_1, \dots, v_{|V|}\}$.

\subsubsection{Multiple Node Selection}

If we desire to perturb $k$ nodes, a simple solution is to choose the node having the best poisoning efficiency from the rest of candidates $k$ times.

\subsection{Extension 2: Suppression of Infections}
\label{sec:ext1}

This is an extension of \method.
Whenever adding perturbations on a node, there is possibilities to propagate the poisoned information through graph convolutions.
We call this unfortunate propagation \textit{infections}.

To mitigate the number of infected nodes, we introduce a penalty into (\ref{eq:L}).
The penalty is defined as:
\begin{equation}
    p(X') = \sum_{q \in Q} \left(\max_{i \neq t}(Z(X', A)_{q,i}) - Z(X',A)_{q,c}\right)_+
\label{eq:penalty}
\end{equation}
where $Q$ is a set of nodes except $u$ and $v$, $c$ is non-deceived output label. 
This is the penalty to add loss if the output label is changed via the perturbations.
Finally, we solve the following objective function:
\begin{equation}
    \mathcal{L}^*(x'_v, u, t, \lambda, \beta) = \mathcal{L}(x'_v, u, t, \lambda) + \beta \cdot p(X').
    \label{eq:obj_ext}
\end{equation}

\section{Evaluation}

This section demonstrates the effectiveness of our proposed attack \method with two datasets.
The experimental evaluations were designed to answer following questions:
\begin{itemize}
\setlength{\parskip}{0cm}
    \item How successful is our method in leading misclassification into node classifiers with GCNs?
    \item How far nodes can our method perform success from?
    \item How successful is our method in choosing the poisoning node which achieve adversarial attack with smaller perturbations?
\end{itemize}

\paragraph{Dataset}
As well as \cite{zugner2018adversarial}, we utilize CORA-ML \cite{mccallum2000automating} and CiteSeer networks as in \cite{sen2008collective}, whose characteristics are described in Table \ref{tbl:dataset}.
We split the network in labeled (20\%) and unlabeled nodes (80\%). 
We further split the labeled nodes in equal parts training and validation sets to train the node classifiers which our attack try to deceive.

\paragraph{Model architectures}
We employ well-known graph convolutional neural networks, GCN with two graph convolutional layers with semi-supervised setting described above \cite{kipf2017semi} \footnote{\url{https://github.com/tkipf/pygcn}}.
In the following evaluations, we also use GCNs with 3 layers and 4 layers.
The detail of the model architectures of the those GCNs are described in Table \ref{tbl:arch}. 
In the training for those GCNs, we set learning rate is 0.01, dropout rate is 0.5. We iterate training within 200 epochs.

\paragraph{Setting of Our Attack}
Our attack \method iteratively searches adversarial perturbations with smaller size of modifications by binary search.
We set max\_search\_steps=9, max\_iter=1000, $\lambda_{init}=1.0$, $\lambda_{max\_init}=10^9$, $\lambda_{min\_init}=0$, where max\_search\_steps represents number of binary search steps, $\lambda_{init}$ represents initial value of $\lambda$.
We developed \method in Python 3.6 and PyTorch 1.0.0.

\paragraph{Competitor}
We employ Nettack's indirect attack\cite{zugner2018adversarial} to compare effectiveness with proposed attack.
The indirect attack automatically chooses given number of influencer nodes, which are nodes around the target.
We set the number of influencer nodes as 1.
Since Nettack perturbs number of features within given budget, we utilize linear search to find positive adversarial examples.
The linear search iteratively increase the budget from 1 while score (\ref{eq:untarget}) is decreased and positive adversarial is not found.

\begin{table}[t]
%\small
%\begin{minipage}{0.4\hsize}
\centering
\begin{tabular}{lcccc}
\toprule
         & \#nodes  & \#edges   & \#features & \#classes  \\
%         & $n$ & $E$   & $d$ & $L$  \\
\midrule
CORA-ML \cite{mccallum2000automating}  & 2708     & 13264     & 1433  & 7  \\
CiteSeer \cite{sen2008collective} & 3312     & 12384     & 3703  & 6  \\
\bottomrule
\end{tabular}
\caption{Dataset}
\label{tbl:dataset}
\end{table}
%\end{minipage}
\begin{table}[t]
%\small
%\hfill
%\begin{minipage}{0.58\hsize}
\centering
\begin{tabular}{lccc}
\toprule
Layer Type          & GCN(2)    & GCN(3)    & GCN(4)    \\
\midrule
GConv + ReLU ($d$)    & \checkmark & \checkmark & \checkmark \\
GConv + ReLU (256)    &   &   & \checkmark \\
GConv + ReLU (64)     &   & \checkmark & \checkmark \\
GConv (16)            & \checkmark & \checkmark & \checkmark \\
Softmax (\#classes)          & \checkmark & \checkmark & \checkmark \\ 
\bottomrule
\end{tabular}
\caption{Model Architectures}
\label{tbl:arch}
%\end{minipage}
\end{table}

\subsection{Attack Success Rate}

\begin{table*}[t!]
\centering
\hfill
\subfloat[CORA-ML]{
    \begin{tabular}{lccccc}
    \toprule
    \multicolumn{2}{c}{} & \multicolumn{4}{c}{Attack Success Rate} \\
    Attack & $f$ & 1-hop  & 2-hop & 3-hop & 4-hop \\ \cmidrule(lr){1-2} \cmidrule(lr){3-6}
    \method & GCN(2) & 1.00   & 0.92  & 0.00  & 0.00  \\ 
            & GCN(3) & --     & --    & 0.54  & 0.00  \\ 
            & GCN(4) & --     & --    & --    & 0.17  \\
    Nettack & GCN(2) & 0.68   & --    & --    & --   \\
    \bottomrule
    \end{tabular}
    \label{tbl:cora}
}
\hfill
\subfloat[CiteSeer]{
    \begin{tabular}{lccccc}
    \toprule
    \multicolumn{2}{c}{} & \multicolumn{4}{c}{Attack Success Rate} \\
    Attack & $f$ & 1-hop  & 2-hop & 3-hop & 4-hop \\ \cmidrule(lr){1-2} \cmidrule(lr){3-6}
    \method & GCN(2) & 1.00 &  1.00  & 0.08  & 0.00  \\
            & GCN(3) & --     & --    & 0.87  & 0.06  \\
            & GCN(4) & --     & --    & --    & 0.64  \\
    Nettack & GCN(2) & 0.55   & --    & --    & --   \\
    \bottomrule
    \end{tabular}
    \label{tbl:citeseer}
}
\hfill
\caption{Overall Attack Success Rate. $m$-layer GCN could be deceived by \method within $m$-hop neighbors.}
\end{table*}

Here we answer the question: How successful is our method in leading misclassification into the target node from its neighbors?

To evaluate the effectiveness of our \method, we measure attack success rate which is the fraction of positive adversarial examples whose size of perturbations are less than threshold.
The attack success rate $success(\theta)$ is defined as follows:
\begin{equation*}
    success(\theta)=\frac{|\{x'_i \in X' | dist(x_i, x'_i)<\theta \}|}{|X|}
\end{equation*}
where $dist(x_i, x'_i)=||x_i - x'_i||_2$.

We measure the attack success rates for each distance of poisoning neighbors.
To measure the attack success rates, we crafted 200 adversarial perturbations through randomly choosing triples of (target node, target class, poisoning node) for each case.
Here we do not employ the proposed poisoning node selection.

\begin{figure}[t!]
    \centering
    \subfloat[CORA-ML]{
        \includegraphics[width=0.485\hsize]{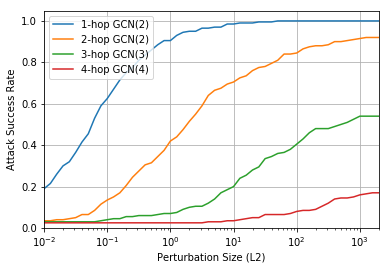}
        \label{fig:cora}
    }
    \subfloat[CiteSeer]{
        \includegraphics[width=0.485\hsize]{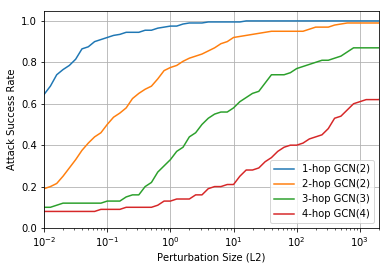}
        \label{fig:citeseer}
    }
    \caption{Attack success rate of our indirect attack from single poisoned neighbor. Adversary can fool any nodes in 2-hops from the hijacked neighbor with high-confidence.}
   \label{fig:preview}
\end{figure}

We show the attack success rate under GCN(2) for Cora-ML and CiteSeer in Figure \ref{fig:preview} and Figure \ref{fig:citeseer} respectively.
On the both figure we plot the attack success rates when \method poisons 1-hop neighbors (blue line) and 2-hop neighbors (yellow line).

\paragraph{Observations}
Figure \ref{fig:preview} shows very high attack success rates for both at 1-hop and 2-hop neighbors.
In 1-hop neighbors, even when the perturbation size is less 1.0(=$10^0$), the attack success rate is more than 90 $\%$.
Poisoning 1-hop neighbors also achieve complete attack success after the perturbation is around 50.
Adversarial perturbations on 2-hop neighbors show 92 $\%$ attack success rate in total.
The fact that we can deceive any nodes' classification results through poisoning single node which have no direct connections between the target is very important.
Figure \ref{fig:citeseer} also shows very high attack success rates.
Both poisoning 1-hop and 2-hop neighbors show complete attack success at the end.
When the perturbations is less than 1.0, 1-hop shows more than 95 $\%$ and 2-hop shows around 80 $\%$ attack success rate respectively. 
Tables \ref{tbl:cora} and \ref{tbl:citeseer} shows overall attack success rate compare with Nettack.
Our attack shows higher success rate than Nettack with 1-influencer setting which iteratively perturbs a feature.

\paragraph{Discussion}
Remarkable thing here is two-layers GCN can be deceived classification results from poisoning nodes at 2-hops far from the target node.
We can say that GCNs are vulnerable not only modifications of directly connecting neighbors but nodes at 2-hops far.
In social networks, 2-hops neighbors are friends of friends.
Most of them are unknown instances that we do not care about.
Thus, it is very hard to notice about becoming a victim.
Graph convolutional neural networks are very powerful machine learning tools, but we need to consider risks against adversarial perturbations.

\subsection{Perturbations on Remote Nodes}

Beyond 2-hops from the target, can \method generate positive adversarial perturbations?

\paragraph{Observations}
Table \ref{tbl:cora} and \ref{tbl:citeseer} demonstrate the attack success rates of poisoning 2-hop neighbors under GCN(2), 3-hop neighbors under GCN(3) and 4-hop neighbors under GCN(4).
On CORA-ML, the success rate at 3-hops under GCN(3) is more than 50 $\%$ while 0 under GCN(2).
Similarly, the success rate at 4-hops under GCN(4) is 17 $\%$ while 0 under GCN(2) and GCN(3).
Fig \ref{fig:preview} and \ref{fig:citeseer} also demonstrate the attack success rate along with the perturbation size.
Positive adversarials at 3-hops consumed much more perturbations than 2-hops. 
Similarly, positive adversarials at 4-hops consumed lots of perturbations as well.

\paragraph{Discussions}
Based on the above experimental results, we can say that it is possible to craft possible adversarials at nodes far from the target even when we built multi-layer GCN, but \method cannot craft positive adversarials in high-confidence.
We demonstrated that $m$-layer GCN could be deceived within $m$-hop neighbors.

\subsection{Effectiveness of Poisoning Node Selection}

\begin{figure}[t!]
    \centering
    \subfloat[CORA-ML]{
        \includegraphics[width=0.485\hsize]{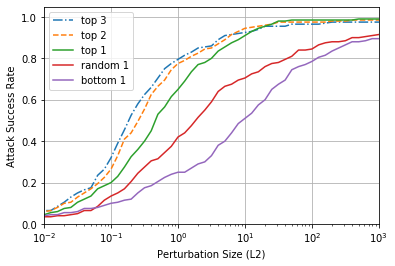}
        \label{fig:infl_cora}
    }
    \subfloat[CiteSeer]{
        \includegraphics[width=0.485\hsize]{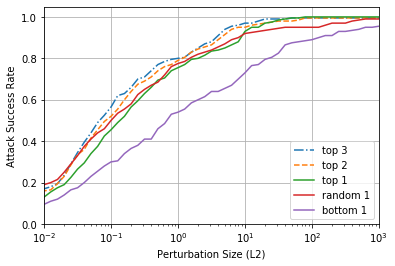}
        \label{fig:infl_citeseer}
    }
    \caption{Our poisoning node selection is effective. \method with perturbing the node having the highest poisoning efficiency shows significantly higher attack success rate.}
    \label{fig:infl}
\end{figure}

Here we answer the question: How successful is our method in choosing the poisoning node which achieve adversarial attack with smaller perturbations?

To evaluate the effectiveness of our poisoning node selection, we measure the attack success rate $success(\theta)$ and the average size of perturbations.
We compare the attacks which perturb top-1 node in poisoning efficiency, top-2 nodes, top-3 nodes, bottom-1 and random from 2-hop neighbors around each target.
Result of top-k is made by the attacks perturbing nodes having top-k scores. 
Random is identical to the result in Fig \ref{fig:preview}.
In this evaluation, we crafted 200 adversarial perturbations through randomly choosing (target node, target class) for each case as well.

\paragraph{Observation 1 (Success Rate with Node Selection)}
In Figure \ref{fig:infl_cora}, \method with choosing top-1 outperforms the one with random selection and bottom-1.
At the perturbation size is 1.0, \method with top-1 shows around 65 \% success rate.
It is higher than \method with random selection, which shows around 40 \% in Figure \ref{fig:preview}.
Figure \ref{fig:infl_citeseer} also shows very high attack success rates.
The differences between the top-1 and the bottom-1 are also large.
In Figure \ref{fig:infl_cora} and \ref{fig:infl_citeseer}, \method with poisoning top-2 and top-3 nodes outperform the top-1.
Those attacks enable us to craft higher confidence attack at a level of perturbation.
At the perturbation size is 1.0, \method with top-3 nodes shows over 80 \% success rate for both two data.

\begin{figure}[t!]
    \centering
    \includegraphics[width=.9\hsize]{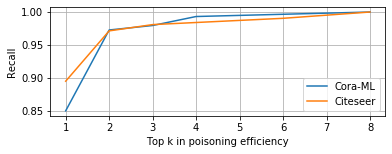}
    \caption{Effectively discovered the smallest perturbation node.}
    \label{fig:recall_topk}
\end{figure}

Next, to check the effectiveness of our poisoning node selection, we measure recall in finding the node which give us the smallest perturbation.
Figure \ref{fig:recall_topk} shows the recall at top k highest poisoning efficiency nodes.
In this evaluation, we randomly pick 200 target nodes.
For each target, we compare all attacks which perturb single node having different poisoning efficiency in 2-hop neighbors.
We attempt all attacks using single poisoning node $v$ chosen from $\mathcal{P}_u$ that satisfies 
$\mathcal{P}_u = \{ p | \phi_u^{(2)}(p) \neq \phi_u^{(2)}(q) \land p \neq q \land p, q \in \mathcal{N}_u^{(2)} \}$.
Note that, in our pre-study, the nodes sharing the same poisoning efficiency tended to result in very similar size of perturbations.

\paragraph{Observation 2 (Recall in Discovering the Smallest Perturbation Node)}
Figure \ref{fig:recall_topk} shows that 85\% of attacks achieved the smallest perturbations on Cora-ML, and 90\% on Citeseer.
At top-2, our method got more than 96\% recall.% in both data.

We further evaluate rank correlation between ranks in poisoning efficiency and ranks in the size of perturbations.
To measure the rank correlation, we compute Spearman’s rank correlation coefficient \cite{spearman1904proof} defined as follows:
\begin{equation*}
    rcc(u) = \frac{6\sum_{v \in \mathcal{P}_u} d_v^2}{\kappa_u(\kappa_u^2-1)}
\end{equation*}
where $d_v$ is the difference between two ranks of $v \in \mathcal{P}_u$ and $\kappa_u = |\mathcal{P}_u|$.
The output $rcc(u)$ is within [-1,1].
+1 indicates a perfect association, 0 indicates no association, -1 indicates a perfect negative association between two ranks.

\paragraph{Observation 3 (Rank correlation between poisoning efficiency and perturbation size)}
In Table \ref{tbl:rcc}, mean of the rank correlation between the rank of poisoning candidate nodes in poisoning efficiency and in perturbation size are more than 0.9.
Thus, these two ranks have very strong association.
Therefore, our proposed method with the poisoning node selection is effective to craft adversarial attacks with smaller perturbations in many cases.

\begin{table}[t]
\centering
\begin{tabular}{lcc}
\toprule
         & mean  & std \\
\midrule
CORA-ML  & 0.922    & 0.238  \\
CiteSeer & 0.962    & 0.121  \\
\bottomrule
\end{tabular}
\caption{Rank correlation coefficient. Ranks in poisoning efficiency and in perturbation sizes shows strong associations.}
\label{tbl:rcc}
\end{table}

\paragraph{Discussions}
From the above results, our method can successfully choose the poisoning node which needs small perturbations to deceive the node classifier.
The proposed poisoning node selection is a heuristic way with considering how much information can be delivered from a candidate to the target.
Our simple, intuitive and light-weight node selection helps \method to achieve high-confident adversarial perturbations with small noises.

\begin{table}[t!]
\centering
%    \small
    \begin{tabular}{llrrrr}
    %\hline
    \toprule
    
    \multicolumn{2}{c}{\method} & median & mean     & std. dev   & zeros\\
    \cmidrule(lr){1-2}
    \cmidrule(lr){3-6}
%                & penalty & median & mean     & std   & zeros   \\ \midrule
    1-hop & $\beta = 0$   & 4      & 11.28    & 23.37 & 0.02    \\ %\hline
     & $\beta = 0.01$    & 4      & \textbf{9.31} & \textbf{17.66} & \textbf{0.04}  \\ %\hline%\hline
    \cmidrule(lr){1-2}
    \cmidrule(lr){3-6}
    2-hop & $\beta = 0$   & 13.5   & 22.24    & 43.90  & 0.00   \\ %\hline
     & $\beta = 0.01$    & \textbf{12.5} & \textbf{22.11} & \textbf{43.85}  & 0.00   \\
     \bottomrule
    \end{tabular}
\caption{Infection Nodes}
\label{tbl:infec}
\end{table}

\begin{table*}[t]
\centering
\hfill
\subfloat[CORA-ML]{
    \begin{tabular}{lrrr}
    %\hline
    \toprule
    poisoning nodes & success rate & mean L2 loss & mean \#infec. \\
    \cmidrule(lr){1-1}
    \cmidrule(lr){2-4}
    top 1  & 0.990 & 4.769  & 12.010 \\
    top 2  & 0.985 & 4.238  & 12.871 \\
    top 3  & 0.980 & 17.982 & 19.148 \\
    bottom 1    & 0.895 & 44.679 & 23.078 \\ 
     \bottomrule
    \end{tabular}
    \label{tbl:infl_cora}
}
\hfill
\subfloat[CiteSeer]{
    \begin{tabular}{lrrr}
    %\hline
    \toprule
    poisoning nodes & success rate & mean L2 loss  & mean \#infec. \\
    \cmidrule(lr){1-1}
    \cmidrule(lr){2-4}
    top 1    & 1.000    & 2.665    & 7.080 \\
    top 2   &  0.995    & 2.114    & 9.603 \\
    top 3   &  0.995    & 1.204    & 11.312 \\
    bottom 1 & 0.970    & 48.254   & 12.871 \\ 
     \bottomrule
    \end{tabular}
    \label{tbl:infl_citeseer}
}
\hfill
\caption{\method with the poisoning node selection shows higher attack success rate and smaller perturbations. Number of infection nodes is increased when \method perturb multiple nodes.}
\end{table*}

\subsection{Infections}

We evaluate how many nodes are infected by the adversarial perturbations.
In the adversary's view, smaller number of infected nodes is better to conceal her malicious activity.

Table \ref{tbl:infec} shows statistics about number of infected nodes when adding perturbations on 1-hop or 2-hop nodes.
We also measure the statistics with and without the penalty introduced in (\ref{eq:penalty}).
In case we turn on the penalty, we set $\beta=0.01$ on (\ref{eq:obj_ext}).
While, $\beta=0$ represents proposed method with no penalty.
We evaluate number of infections for CORA-ML dataset.
Here we do not mention about CiteSeer dataset because the results had very small number of infections.

First we look at the results without the penalty.
Adversarial attacks from 1-hop neighbors infect a few nodes.
Attacks from 2-hop neighbors turn much more number of nodes into wrong results than 1-hops.
Next, we proceed into the \method with the penalty.
For both 1-hop and 2-hop, the number of infection nodes are decreased.
In 1-hop, the number of zero infections (that is zeros in Table \ref{tbl:infec}) are increased.

Since perturbation size of 2-hop neighbors' attacks is larger than 1-hop, the number of infection nodes is also increased.
Fortunately, \method with the penalty can mitigate the infections.
However, it have not revealed significant benefits yet.
Finding more effective value of $\beta$ needs more studies.

Table \ref{tbl:infl_cora} and Table \ref{tbl:infl_citeseer} show the average size of perturbations (L2 loss), overall attack success rate, and number of infections, for \method with poisoning different types of nodes.
Those success rates are higher than \method with random choices (Table \ref{tbl:infl_cora} and \ref{tbl:infl_citeseer})
The size of perturbations of top-1 is more than 10 times smaller than bottom-1 for each data.
When we perturb several nodes' features, \method has much more chance to reduce the size of perturbations because it may have more freedom to perturb.
However, number of infections are increased (Table \ref{tbl:infl_cora} and \ref{tbl:infl_citeseer}).

\section{Conclusion}

Towards evaluating adversary robustness of GCNs, we tackled the question; can we generate effective adversarial perturbations on a node far from the target?
We introduced a new attack named \method which poisons a node's features to lead misclassification into a target far more than one-hop. 
We also introduced an approach to discover the poisoning node with smaller perturbations.
In our evaluations, attack success rates of the proposed attack were at most 100\% from 1-hop neighbors and 92\% from two-hop neighbors in Cora-ML dataset by poisoning single randomly selected node.
The proposed attack can be used as a benchmark in future defense attempts to develop graph convolutional neural networks with robustness against indirect adversarial perturbations.

%\balance

\bibliographystyle{abbrv}
\bibliography{ref_bigdata19}

\end{document}